\documentclass[a4paper]{article}

\usepackage{authblk}
\usepackage[utf8]{inputenc}
\usepackage[T1]{fontenc}
\usepackage{pslatex}
\usepackage[english]{babel}
\usepackage{fancyhdr}
\usepackage{hyperref}
\usepackage{multirow}
\usepackage{amsmath,amssymb,amsfonts}
\usepackage{graphicx}
\usepackage{textcomp}
\usepackage{xcolor}
\usepackage{algorithm}
\usepackage[noend]{algpseudocode}
\usepackage{url}
\usepackage{hyperref}
\usepackage{microtype}
\usepackage{subcaption}
\usepackage{tikz}
\usetikzlibrary{positioning}
\usetikzlibrary{decorations.pathreplacing}
\usepackage{cite}
\usepackage{wasysym}
\usepackage{booktabs}
\usepackage{paralist}
\usepackage{hyperref}

\newcommand{\F}{\mathbb{F}}

\providecommand{\keywords}[1]{\textbf{\textit{Keywords }} #1}

\begin{document}

\title{On the Evolution of Boomerang Uniformity in Cryptographic S-boxes}

\author[1]{Marko Djurasevic}
\author[1]{Domagoj Jakobovic}
\author[2]{Luca Mariot}
\author[3]{Sihem Mesnager}
\author[2]{Stjepan Picek}

\affil[1]{{\normalsize University of Zagreb, 10000 Zagreb, Croatia} \\

    {\small \texttt{\{marko.durasevic, domagoj.jakobovic\}@fer.hr}}}

\affil[2]{{\normalsize Digital Security Group, Radboud University, Postbus 9010, 6500 GL Nijmegen, The Netherlands} \\
	
	{\small \texttt{\{luca.mariot,stjepan.picek\}@ru.nl}}}

\affil[3]{{\normalsize Department of Mathematics, University of Paris VIII, F-93526 Saint-Denis, France} \\
	
	{\small \texttt{sihem.mesnager@gmail.com}}}
	
\maketitle

\begin{abstract}
\noindent
S-boxes are an important primitive that help cryptographic algorithms to be resilient against various attacks. The resilience against specific attacks can be connected with a certain property of an S-box, and the better the property value, the more secure the algorithm. One example of such a property is called boomerang uniformity, which helps to be resilient against boomerang attacks. How to construct S-boxes with good boomerang uniformity is not always clear. There are algebraic techniques that can result in good boomerang uniformity, but the results are still rare. In this work, we explore the evolution of S-boxes with good values of boomerang uniformity. We consider three different encodings and five S-box sizes. For sizes $4\times 4$ and $5\times 5$, we manage to obtain optimal solutions. For $6\times 6$, we obtain optimal boomerang uniformity for the non-APN function. For larger sizes, the results indicate the problem to be very difficult (even more difficult than evolving differential uniformity, which can be considered a well-researched problem).
\end{abstract}


\section{Introduction}
\label{sec:introduction}

S-boxes (Substitution boxes, $(n,m)$ functions) are mathematical objects commonly used in block ciphers to provide resilience against various attacks~\cite{carlet_2021}.
To do so, S-boxes need to be carefully selected so that their cryptographic properties allow resilience against attacks. 
Today, the most common option is to use S-boxes with $n=m$. Since the search space of all S-boxes is huge ($2^{n 2^n}$), exhaustive search is not easy already for S-boxes of size $4\times 4$, and for larger sizes, it is impossible.
Thus, constructing S-boxes with good cryptographic properties has been an active research domain for more than 30 years on. The common options are to use 1) algebraic constructions, 2) random search, and 3) metaheuristics. Among these, algebraic constructions are the most accepted option as it guarantees the quality of the attainable cryptographic properties. 

Nevertheless, not all constructions will achieve optimal values for all relevant cryptographic properties. Luckily, this is not a problem for well-known properties like nonlinearity and differential uniformity, as we know a number of constructions that will give excellent (if not optimal) properties. Still, as community knowledge progresses, new properties are developed, and for some of those, it becomes less clear what algebraic construction to use (if any).

One example of such a property is the boomerang uniformity~\cite{Boura_Canteaut_2018}, which has been proposed only recently. This property is not preserved under all notions of equivalence (discussed in Section~\ref{sec:background}). %This means that S-boxes that have, e.g., the same differential uniformity can have different boomerang uniformity. Even more, it 
This means that we cannot ``count'' on every algebraic construction to provide an S-box with the desired boomerang uniformity, and consequently, it makes sense to investigate different ways how to construct S-boxes with good boomerang uniformity. In this paper, we use evolutionary algorithms (EAs) for this goal.
Unfortunately, it is far from trivial to construct S-boxes with good boomerang uniformity. We already discussed one reason: the search space size. The other difficulty lies in the computational cost of evaluating the boomerang uniformity, which is $\mathcal O(2^{3n})$.

At the same time, it is relevant to investigate how well evolutionary algorithms can perform for this problem. Since we know the best possible values for boomerang uniformity for different S-box sizes, we can easily determine how well EAs perform. Thus, we can consider it as a benchmark problem for EAs.
Additionally, if we manage to find (many) solutions with optimal (or excellent) boomerang uniformity, some of those solutions could lead to new insights into which algebraic constructions should be used when requiring S-boxes with good boomerang uniformity.

To the best of our knowledge, there are no related works that consider how metaheuristics perform for the problem of constructing S-boxes with good boomerang uniformity. Still, a number of related works (a subset of those works is discussed in Section~\ref{sec:related} indicate that EAs can evolve S-boxes with good cryptographic properties, especially for smaller sizes (e.g., $4\times 4$ should be easy while $8 \times 8$ is impossible).

In this work, we consider three different encodings to evolve S-boxes: integer, permutation, and CA-based encoding.
Our main contributions are:
\begin{compactenum}
\item We are the first to consider the evolutionary algorithm approach to construct S-boxes with good boomerang uniformity.
\item We employ random search as the baseline technique and observe that for the smallest S-box size ($4\times 4$), the problem is easy.
\item For a number of sizes, we obtain optimal values for boomerang uniformity. More precisely, for $4\times 4$ and $5\times 5$, we reach optimal boomerang uniformity, and for $6\times 6$, optimal boomerang uniformity for the non-APN function.
\item We optimize differential uniformity and boomerang uniformity together with a multi-objective optimization approach, where the results indicate that boomerang uniformity could be an even more difficult property to evolve than differential uniformity.
\item Based on the obtained results, we conclude that the CA-based encoding is the most effective as it gives the best results for smaller sizes, while none of the encodings work particularly well for larger S-box sizes.
\end{compactenum}

\section{Background}
\label{sec:background}

This section starts with a brief introduction to the notation we follow. Afterward, we provide relevant information about (bijective) S-boxes, equivalence relations, differential uniformity, and boomerang uniformity.

\subsection{Notation}

Let $n, m$ be positive integers, i.e.,  $n, m \in \mathbb{N}^+$, and let $\mathbb{F}_{2}$ be the Galois field (GF) with two elements. By $\mathbb{F}_{2}^{n}$ and $\mathbb{F}_{2^n}$ we denote respectively the $n$-dimensional vector space over $\mathbb{F}_{2}$ and the field extension of $\F_2$ with $2^n$ elements. The addition of elements in $\F_2^n$ and $\mathbb{F}_{2^n}$ are denoted respectively with ``+'' and ``$\oplus$''.
%A derivative of a function $F$ in the direction of $a$ is denoted as $D_aF$ and it equals $F(x)+F(x+a).$

\subsection{S-boxes}

An S-box (substitution box, or $(n, m)$-function) is a mapping $F$ from $n$ bits into $m$ bits.  
An $(n, m)$-function $F$ can be defined as a vector $F = (f_1,\cdots,f_m)$, where the Boolean functions $f_i: \mathbb{F}_2^n \rightarrow \mathbb{F}_2$ for $i \in \{1, \cdots, m\}$ are called the coordinate functions of $F$. %The component functions of an $(n, m)$\nobreakdash-\hspace{0pt}function $F$ are all the linear combinations of the coordinate functions with non all-zero coefficients.
As for every $n$, there exists a field  $\mathbb{F}_{2^n}$ of order $2^n$, and we can endow the vector space $\mathbb{F}_2^n$ with the structure of that field when convenient. 

\subsubsection{Equivalence Relations}

A function $F: \mathbb F_2^n \rightarrow \mathbb F_2^n$ is linear if:
\begin{equation}
\label{eq:linear}
    F(x) = \sum_{0 \leq i < n} a_i x^{2^i}, a_i \in \mathbb F_{2^n}.
\end{equation}
A function $F$ is affine if it is a sum of a linear function (Eq.~\eqref{eq:linear}) and a constant term.

Two functions $F: \mathbb F_{2^n} \rightarrow \mathbb F_{2^m}$ and $G: \mathbb F_{2^n} \rightarrow \mathbb F_{2^m}$ are called:
\begin{compactenum}
    \item Affine equivalent if $G = A \circ F \circ B$, where the mappings $A$ and $B$ are affine permutations on $\mathbb F_{2^m}$ and $\mathbb F_{2^n}$, respectively.
    \item Extended affine equivalent (EA-equivalent) if $G = A \circ F  \circ B + C$, where $A$ and $B$ are affine permutations on $\mathbb F_{2^m}$ and $\mathbb F_{2^n}$, respectively, and where $C$ is an affine function from $\mathbb F_{2^n}$ to $\mathbb F_{2^m}$.
    \item Carlet-Charpin-Zinoviev equivalent (CCZ-equivalent) if for an affine permutation $\mathcal A$ of $\mathbb F_{2^n} \times \mathbb F_{2^m}$, the image of the graph of $F$ is the graph of $G$, i.e.: $\mathcal{A}(\left\lbrace (x, F(x)), x \in \mathbb F_{2^n} \right\rbrace ) = \left\lbrace (x, G(x)), x \in \mathbb F_{2^n} \right\rbrace$~\cite{10.1023/A:1008344232130}.
\end{compactenum}
A cryptographic property is called invariant if it is preserved by a certain equivalence notion.

\subsubsection{Balancedness}

An $(n, m)$-function $F$ is balanced if it takes every value of $\mathbb{F}_{2}^{m}$ the same number $2^{n - m}$ of times.
For an S-box to be balanced, it needs to be a permutation. In the rest of this work, we will concentrate on bijective S-boxes, i.e., those where the input and output dimensions are the same ($n=m)$.

\subsubsection{Differential Uniformity and APN Functions}

Let $F$ be a function from $\mathbb{F}_2^n$ into $\mathbb{F}_2^m$ with $a \in \mathbb{F}_2^n$ and $b \in \mathbb{F}_2^m$, then:
\begin{equation}
\label{eq:diff}
D_F (a, b) =  \left\lbrace x \in \mathbb{F}_2^n : F(x)+F(x+a) =b\right\rbrace.
\end{equation}
The entry at the position $(a, b)$ corresponds to the cardinality of the delta difference table $D_F (a, b)$ and is denoted as $\delta (a, b)$. The differential uniformity $\delta$ is defined as~\cite{kaisa}:
\begin{equation}
\label{eq:delta}
\delta = \max_{\substack{a \neq 0, b}} \delta (a, b).
\end{equation}

The differential uniformity must be even since the solutions of Eq.~\eqref{eq:diff} go in pairs. Indeed, if $x$ is a solution of $F(x) + F(x + a) = b$, then $x + a$ is
also a solution. The lower the differential uniformity, the better the S-box contribution to withstand the differential attack~\cite{carlet_2021}.
When the number of output bits is the same as the number of input bits, differential uniformity is equal to or greater than 2.
Functions with differential uniformity equal to 2 are called Almost Perfect Nonlinear (APN) functions.
APN functions exist for both odd and even numbers of variables. 
When discussing the differential uniformity parameter for permutations, the best possible (and known) value is 2 for any odd $n$ and $n = 6$. For $n$ even and larger than 6, this is an open question. 
For $n=4$, the best possible differential uniformity equals 4, and for $n=8$, the best-known value equals 4.
Differential uniformity is invariant for all previously discussed equivalence relations.

\subsubsection{Boomerang Uniformity}

In 1999, Wagner introduced the boomerang attack, a cryptanalysis technique against block ciphers involving S-boxes~\cite{Wagner1999}.
Then, in 2018, Cid et al. introduced the concept of the Boomerang Connectivity Table (BCT) of a permutation $F$~\cite{Cid2018}.
The same year, Boura and Canteaut introduced a property of cryptographic S-boxes called boomerang uniformity which is defined as the maximum value in the BCT~\cite{Boura_Canteaut_2018}.

Let $F$ be a permutation over $\mathbb F_{2^n}$, and $a, b \in \mathbb F_{2^n}$. Then, the Boomerang Connectivity Table (BCT) of $F$ is given by a $2^n \times 2^n$ table $T_f$:
\begin{equation}
    \label{eq:bmt}
    T_F(a,b) = |\{x \in \mathbb F_{2^n}: F^{-1}(F(x)+a)+F^{-1}(F(x+b)+a) = b \}|.
\end{equation}

The entry at the position $(a, b)$ corresponds to the cardinality of the boomerang uniformity table $T_F (a, b)$ and is denoted as $\beta (a, b)$. The boomerang uniformity $\beta$ is defined as~\cite{Boura_Canteaut_2018}:
\begin{equation}
\label{eq:boomerang}
\beta = \max_{\substack{a,b \neq 0}} \beta (a, b).
\end{equation}
The boomerang uniformity is invariant for affine equivalence but not for extended affine and CCZ-equivalence~\cite{Boura_Canteaut_2018}.
It has been proved that $\delta \leq \beta$ for any function~$F$~\cite{Cid2018}. Additionally, $\delta = 2$ if and only if $\beta = 2$. 
Moreover, for $n=4$, the lowest boomerang uniformity that can be achieved is 6. For the mapping $F(x)=x^{2^n-2}$, the boomerang uniformity equals 6 if $n \equiv \  0 \ mod \ 4$ or 4 if $n \equiv \ 2 \ mod \ 4$.
Almost all permutations with optimal boomerang uniformity known today are extended-affine equivalent to the Gold function ($F(x)=x^d$, where $d=2^i+1$ and $gcd(i,n)=1$)~\cite{Tian2020}.
More results about boomerang uniformity, especially for quadratic permutations, can be found in~\cite{Mesnager2020}.
%Thus, APN permutations offer optimal resistance to differential and boomerang attacks.

\section{Related Work}
\label{sec:related}

As discussed in Section~\ref{sec:introduction}, metaheuristics are one of the options for designing S-boxes with good cryptographic properties.
In 2004, Clark et al. were the first to use metaheuristics for the design of S-boxes~\cite{1331078}. There, the authors used the principles from the evolutionary design of Boolean functions to evolve S-boxes with sizes up to $8\times 8$ and good nonlinearity. 
P. Tesar used a genetic algorithm and a total tree searching to evolve $8\times 8$ S-boxes with nonlinearity equal up to 104~\cite{tesar}.
Kazymyrov et al. used gradient descent to evolve S-boxes with good cryptographic properties~\cite{cryptoeprint:2013:578}. The approach used here differs from previous works as the authors started with S-boxes with low differential uniformity, and they conducted several steps until they found an S-box with good nonlinearity.
Picek et al. investigated the performance of CGP and GP to evolve $3\times 3$ and $4\times 4$ S-boxes~\cite{10.1145/2739482.2764698}. The authors considered the nonlinearity and differential uniformity properties.

Picek, Rotim, and Cupic developed a cost function capable of reaching high nonlinearity values for several S-box sizes~\cite{rotim}. While optimizing nonlinearity only, the authors also reported differential uniformity.
Picek and Jakobovic used genetic programming to evolve constructions resulting in S-boxes with good cryptographic properties~\cite{10.1145/3319619.3322040}. The authors used a single-objective approach and considered differential uniformity.
Picek et al. used genetic programming to evolve cellular automata rules that can be used to generate S-boxes~\cite{10.1145/3067695.3076084}. The obtained results outperformed other metaheuristic techniques for sizes $5\times 5$ up to $7\times 7$. These results still represent state-of-the-art results obtained with metaheuristics.

These works show that metaheuristics can be used to construct small S-boxes with good cryptographic properties. Still, as the properties considered are CCZ-invariant, a simpler approach would be to use an algebraic construction (that is known to give optimal cryptographic properties) and then transform the S-box into an equivalent one (if required). 

Considering the cryptographic properties that are not invariant, some works consider the side-channel resilience of S-boxes or their implementation properties.
For instance, Ege et al. considered the confusion coefficient property~\cite{7169071} and Picek et al. the transparency order property~\cite{10.1145/2556315.2556319}.
Both properties are relevant to make the S-box more resilient against side-channel attacks.
Picek et al. used evolutionary algorithms to evolve S-boxes that are power- or area-efficient~\cite{10.1007/978-3-319-69453-5_9}.  
Similarly, Picek et al. used genetic programming to find cellular automata rules that result in S-boxes with good implementation properties like latency, area, and power~\cite{10.1145/3075564.3079069}.

\section{Experimental Setup}
\label{sec:setup}

In this section, we describe the solution representation and search algorithms used to optimize the boomerang uniformity in S-boxes.

\subsection{Encodings}

S-boxes can be represented in a number of ways that differ substantially from one another. 
However, the previous related work (as discussed in Section~\ref{sec:related}) shows that no single representation is dominant; often, different representations offer the best results depending on both the size of the S-box and the optimized criteria.
In this work, we employ three encodings to represent S-boxes: integer, permutation, and cellular automata-based encoding.

\subsubsection{Integer encoding}

The simplest encoding represents an $n \times n$ S-box as a vector of integer values of size $2^n$; each element of the vector encodes the S-box output for a corresponding S-box input.
Since an $n \times n$ S-box has $n$ output Boolean values, every element in the vector assumes values in $[0, \dots, 2^n-1]$.

For integer encoding, we define the following operators; there is a single mutation operator that selects a random gene and modifies its value in the defined range $[0, \dots, 2^n-1]$ with uniform distribution.
Crossover is performed using one of three operators: single-point and two-point crossover function by combining a child genotype from two parents (individuals) using either a single or two break points when copying genes to the child.
Finally, average crossover creates the child individual where each gene is a (rounded) mean value between corresponding parent genes.

Note that this representation and operators do not preserve balancedness since every possible output is obtainable for every input; in this case, we first enforce the balancedness property with a penalty term in the fitness function (next subsection) and then optimize for boomerang uniformity.

\subsubsection{Permutation encoding}

Probably the most natural way to represent a \textit{balanced} S-box is the permutation encoding. In this case, the individual is encoded with a permutation of size $2^n$ with elements in the range $[0, \dots, 2^n-1]$. 
This representation preserves the balancedness property.

Genetic operators used for this encoding were as follows; for mutation, three operators are used: insert mutation, inversion mutation, and swap mutation \cite{Vlasic2019}.
As for the crossover operators, we use partially mapped crossover (PMX), position-based crossover (PBX), order crossover (OX), uniform-like crossover (ULX), and cyclic crossover \cite{Vlasic2019}.

\subsubsection{Cellular Automata with Genetic Programming}

The third representation uses the fact that an S-box could be represented as a cellular automaton (CA), with transitions from the input bits as the current state to the output bits as the following state.
The transitions of the cellular automaton can be defined by using a \textit{local update rule}, which is simply a Boolean function of at most $n$ bits with a single output bit.
The CA local rule defines the next state of a given bit $c_i(t+1)$, based on the current state of the same bit and its adjacent bits: $c_i(t)$, $c_{i+1}(t)$, $c_{i+2}(t)$, etc.
The same approach is used in the design of some existing S-boxes, e.g., in  Keccak~\cite{keccak}.

In this case, the S-box is represented with a Boolean function that embodies a CA local rule.
To evolve a suitable local rule, we employ genetic programming and encode the Boolean function as a tree.
The $n$ input Boolean variables of the S-box are used as GP terminals.
The GP uses the function set consisting of several Boolean primitives: NOT (inverting its argument), XOR, AND, OR, NAND, and XNOR, each of which takes two input arguments.
Finally, we use the function IF, which takes three arguments and returns the second one if the first one evaluates to $true$, and the third one otherwise. 
An individual obtained with GP is evaluated in the following manner: all the possible $2^n$ input states are considered, and for each state, the same rule (the evolved Boolean function) is applied in parallel to each of the bits to determine the next state (S-box output).
As in the integer encoding, this representation also does not preserve balancedness, which has to be enforced with a suitable fitness function.

The genetic operators are simple tree crossover, uniform crossover, size fair, one-point, and context preserving crossover~\cite{GPfieldguide} (selected at random), and subtree mutation.
Previous experiments using GP suggest that having a maximum tree depth equal to the size of the S-box is sufficient (i.e., maximum tree depth equals $n$, which is also the number of terminals).

\subsection{Fitness Functions}

Since in all experiments we optimize boomerang uniformity, the first fitness function minimizes the property value:
\begin{equation}
    fitness_1 = \beta.
\end{equation}

The above fitness function is used with permutation encoding; for the other two representations, we add a penalty term that measures the distance from a balanced S-box, which is expressed simply as the number of missing output values, denoted as $BAL$.
Only if this value equals zero the boomerang uniformity property is calculated and used in the minimization of the following fitness function:
\begin{equation}
    fitness_2 = 
\begin{cases}
    2^n + BAL   & \text{if } BAL > 0 \\
    \beta,              & \text{otherwise.}
\end{cases}
\end{equation}

\subsection{Algorithms and Parameters}

In the single-objective optimization, we optimize the boomerang property of the S-box only.
The search algorithm, used with all encodings, is a steady-state evolutionary algorithm with tournament selection. In each iteration, three individuals from the population are selected at random, and the worst of the three selected individuals is eliminated.
A new individual is formed using crossover on the remaining two from the tournament, and the new individual is mutated with a given individual mutation probability.
When either crossover or mutation is applied, only one of the available operators for that encoding is used, selected at random.
Furthermore, we use a random search algorithm as a baseline.

Besides the single-objective, we also employ the multi-objective optimization in which both the boomerang and delta uniformity are optimized; both properties are minimized.
In this part of the experiment, we use the well-known NSGA-II multi-objective evolutionary algorithm~\cite{DBLP:journals/tec/DebAPM02}.%\todo{ref}

All encodings and algorithms are applied with the same population size of 500 individuals.
The individual mutation probability is 0.7. All algorithms use the same stopping criterion, which is 500\,000 function evaluations. These values were selected based on preliminary tuning results.

\section{Experimental Results}
\label{sec:experiments}

In this section, we outline the experimental results, first by examining the performance of each encoding independently and then by comparing the best solutions of all encodings. 
Finally, we also investigate the possibility of simultaneously optimizing the boomerang uniformity and differential uniformity using a multi-objective evolutionary algorithm. %\todo{why is this not given as fitness 3?}
Our experiments consider sizes from $4\times 4$ to $8 \times 8$, as those sizes have the most practical relevance (being used in most of the modern block ciphers that are substitution-permutation networks, where bijectivity of S-boxes is mandatory).

\subsection{Integer Encoding}

Table~\ref{tbl:intres} outlines the results obtained using the integer encoding. 
The results show that using a random search, it was impossible to obtain solutions with a boomerang uniformity for any S-box size except the smallest one since the obtained solutions were not balanced.
The results are denoted with '-' in the table for all such cases. 
On the other hand, when using the evolutionary algorithm, the results show that quite stable results are obtained, usually with only a small dispersion among the individual executions. 
Even as the sizes of the S-boxes increase, the distribution of the solutions is still compact.
 
\begin{table}
\setlength{\tabcolsep}{5pt}
\centering
\caption{Results obtained for the integer encoding}
\label{tbl:intres}
\begin{tabular}{@{}crrrrrr@{}}
\toprule
  \multicolumn{1}{c}{\multirow{2}{*}{S-box size}} & \multicolumn{3}{c}{EA}                                                  & \multicolumn{3}{c}{RS}                                                      \\ \cmidrule(lr){2-4}  \cmidrule(lr){5-7}
                          \multicolumn{1}{c}{}                            & \multicolumn{1}{c}{Min.} & \multicolumn{1}{c}{Avg.} & \multicolumn{1}{c}{Std.} & \multicolumn{1}{c}{Min.} & \multicolumn{1}{c}{Avg.} & \multicolumn{1}{c}{Std.} \\ \midrule
$4\times 4$                 & 6                        & 7.2                      & 1.35                     &        8                        & 14.8                     & 3.35                        \\
                            $5\times 5$                 & 10                       & 10.9                     & 1.01                     &     -                        & -                        & -                            \\
                            $6\times 6$                 & 12                       & 13.9                     & 1.11                     &    -                        & -                        & -                           \\
                            $7\times 7$                 & 16                       & 17.2                     & 1.35                     &   -                        & -                        & -                            \\
                            $8\times 8$                 & 18                       & 20                       & 2.29                     &  -                        & -                        & -                              \\ \bottomrule
%                           & $5\times 5$                 & 35                       & 35.8                     & 0.41                     &                          &                          &                          \\
%                           & $6\times 6$                 & 75                       & 76                       & 0.58                     &                          &                          &                          \\
%                           & $7\times 7$                 & 156                      & 158                      & 0.90                     &                          &                          &                          \\
%                           & $8\times 8$                 & 322                      & 326                      & 1.51                     &                          &                          &                          \\ \bottomrule
\end{tabular}
\end{table}
 
\subsection{Permutation Encoding}
Table~\ref{tbl:permres} outlines the results obtained for the permutation encoding. 
It is interesting to observe that for this encoding, both random search and the evolutionary algorithm perform quite similar across all S-box sizes. 
For example, for smaller S-box values, both achieve the same performance since, in every run, they obtained solutions of equal quality. 
On the other hand, for larger sizes, the evolutionary algorithm obtains better solutions than a random search in a few runs and usually achieves better minimum values. 
Nevertheless, the average values of both methods are similar, especially for the largest S-box size, thus indicating a similar performance between both. 
For the two largest S-box sizes, we see that random search always gets stuck in the same solution, whereas the evolutionary algorithm is sometimes able to obtain a solution of slightly better quality. 

\begin{table}
\setlength{\tabcolsep}{5pt}
\centering
\caption{Results obtained for the permutation encoding}
\label{tbl:permres}
\begin{tabular}{@{}crrrrrr@{}}
\toprule
  \multicolumn{1}{c}{\multirow{2}{*}{S-box size}} & \multicolumn{3}{c}{EA}                                                  & \multicolumn{3}{c}{RS}                                                      \\ \cmidrule(lr){2-4}  \cmidrule(lr){5-7}
                          \multicolumn{1}{c}{}                            & \multicolumn{1}{c}{Min.} & \multicolumn{1}{c}{Avg.} & \multicolumn{1}{c}{Std.} & \multicolumn{1}{c}{Min.} & \multicolumn{1}{c}{Avg.} & \multicolumn{1}{c}{Std.} \\ \midrule
 $4\times 4$                 & 6                        & 6                        & 0                        &        6                        & 6                        & 0                         \\
                            $5\times 5$                 & 8                        & 8                        & 0                        &     8                        & 8                        & 0                             \\
                            $6\times 6$                 & 10                       & 10.5                     & 0.86                     &    10                       & 11.93                    & 0.37                         \\
                            $7\times 7$                 & 12                       & 13.6                     & 0.81                     &    14                       & 14                       & 0                           \\
                            $8\times 8$                 & 14                       & 15.9                     & 0.37                     &  16                       & 16                       & 0                            \\  \bottomrule
\end{tabular}
\end{table}

\subsection{CA-based Encoding}

Table~\ref{tbl:cares} shows the results obtained by the CA-based representation.
For this encoding, we again see that for the smaller S-box sizes, both random search and the evolutionary algorithm %\todo{but isn't this GP?} \todo{Yes, but in the algorithms section as far as I see we only mention the SST GA. Therefore i tried to keep it like this. Because GP is basically just another encoding, and the algorithm is the same. But I can change.} 
perform equally well, with random search even achieving a better result on average for S-box of sizes $5\times 5$, since random search always found the best-obtained value.
However, as the size of the S-boxes increases, the performance of random search significantly deteriorates, and it constantly achieves worse results than the evolutionary algorithm.
Regardless, the results obtained by the evolutionary algorithm can also be seen to deteriorate quite swiftly, which does suggest that this encoding might not be suitable for larger S-box sizes. This is aligned with the results in related works that consider, for instance, differential uniformity~\cite{DBLP:journals/ccds/MariotPLJ19}.

\begin{table}
\setlength{\tabcolsep}{5pt}
\centering
\caption{Results obtained for the CA-based encoding}
\label{tbl:cares}
\begin{tabular}{@{}crrrrrr@{}}
\toprule
  \multicolumn{1}{c}{\multirow{2}{*}{S-box size}} & \multicolumn{3}{c}{EA}                                                  & \multicolumn{3}{c}{RS}                                                      \\ \cmidrule(lr){2-4}  \cmidrule(lr){5-7}
                          \multicolumn{1}{c}{}                            & \multicolumn{1}{c}{Min.} & \multicolumn{1}{c}{Avg.} & \multicolumn{1}{c}{Std.} & \multicolumn{1}{c}{Min.} & \multicolumn{1}{c}{Avg.} & \multicolumn{1}{c}{Std.} \\ \midrule
                             $4\times 4$                                     & 6                        & 6                        & 0                        &         6                        & 6                        & 0                          \\
                                                $5\times 5$                                     & 2                        & 2.93                     & 2.15                     &       2                        & 2                        & 0                       \\
                                                $6\times 6$                                     & 4                        & 15.5                     & 3.39                     &      12                       & 23.2                     & 4.72                         \\
                                                $7\times 7$                                     & 16                       & 26.1                     & 6.06                     &    28                       & 50.2                     & 7.97                             \\
                                                $8\times 8$                                     & 42                       & 73.9                     & 16.5                     &   112                      & 132                      & 16.76                           \\ \bottomrule
\end{tabular}
\end{table}

\subsection{Representation Comparison}

Table~\ref{tbl:repcomp} outlines the best result achieved by each of the encodings for the tested S-box sizes.
The best value obtained for each size is denoted in bold.
The results show that for the smallest S-box size, all three encodings are equally successful since they all obtain a value of 6 for boomerang uniformity.
For S-box sizes of $5\times 5$ and $6\times 6$, the CA-based encoding clearly achieves the best results, whereas the remaining two encodings achieve similar results, with the permutation encoding performing slightly better.
However, for the two largest S-box sizes that were tested, we observe that the performance of the CA-based encoding is not as good, especially for the S-box size of $8 \times 8$, for which it achieves poor results.
For these two sizes, the best results are obtained by permutation encoding, followed by integer encoding.

\begin{table}
\setlength{\tabcolsep}{5pt}
\centering
\caption{Best results obtained by tested encodings for various S-box sizes}
\label{tbl:repcomp}
\begin{tabular}{crrr}
\hline
S-box size  & \multicolumn{1}{c}{Int} & \multicolumn{1}{c}{Perm} & \multicolumn{1}{c}{CA} \\ \hline
$4\times 4$ & \textbf{6}              & \textbf{6}               & \textbf{6}             \\
$5\times 5$ & 10                      & 8                        & \textbf{2}             \\
$6\times 6$ & 12                      & 10                       & \textbf{4}             \\
$7\times 7$ & 16                      & \textbf{12}              & 16                     \\
$8\times 8$ & 18                      & \textbf{14}              & 42                     \\ \hline
\end{tabular}
\end{table}
 
\begin{figure}[!tbh]
     \centering
     \begin{subfigure}[b]{0.49\textwidth}
         \centering
         \includegraphics[width=\textwidth]{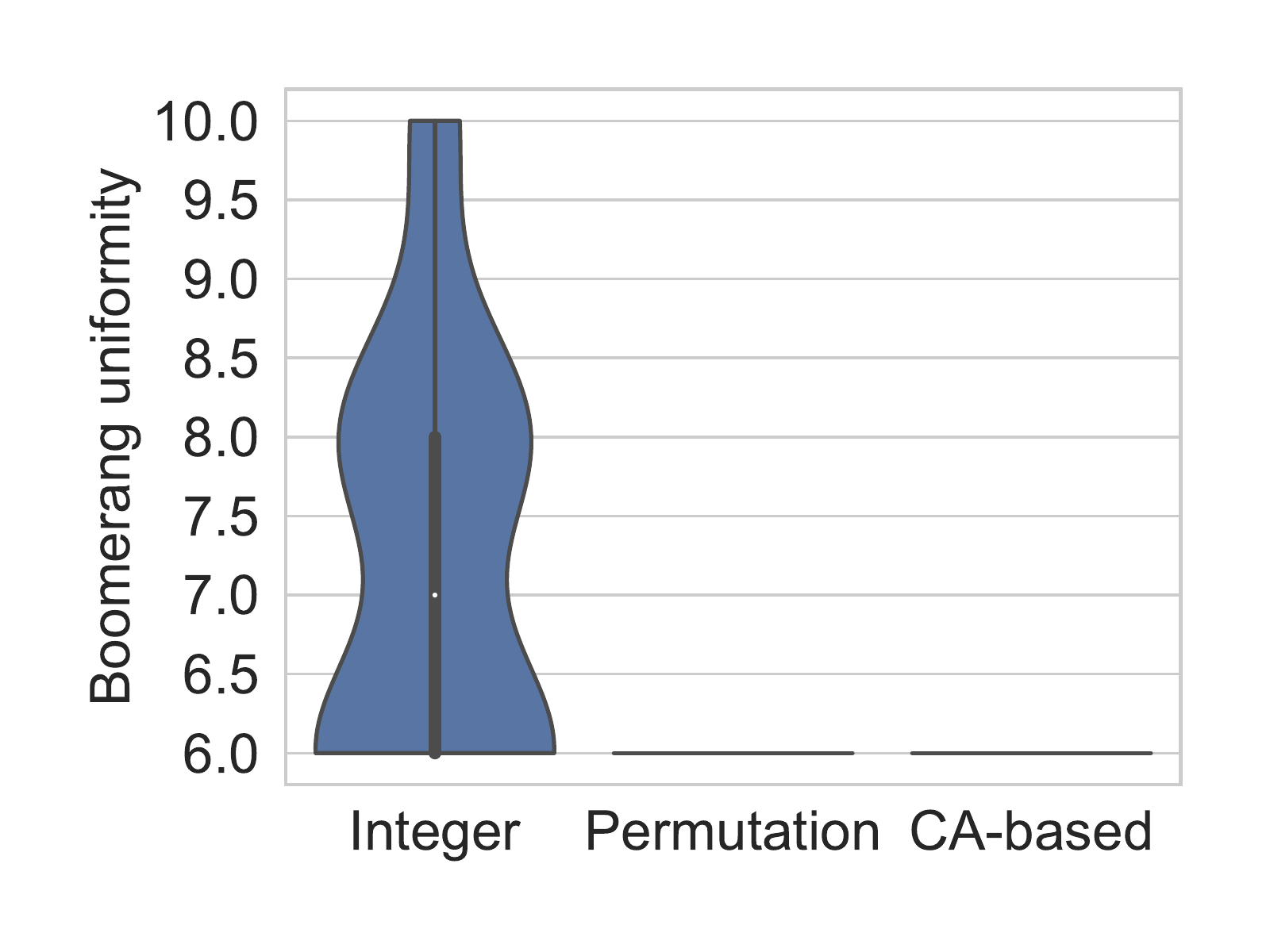}
         \caption{$4\times 4$}
         \label{fig:violin4}
     \end{subfigure}
     \hfill
     \begin{subfigure}[b]{0.49\textwidth}
         \centering
         \includegraphics[width=\textwidth]{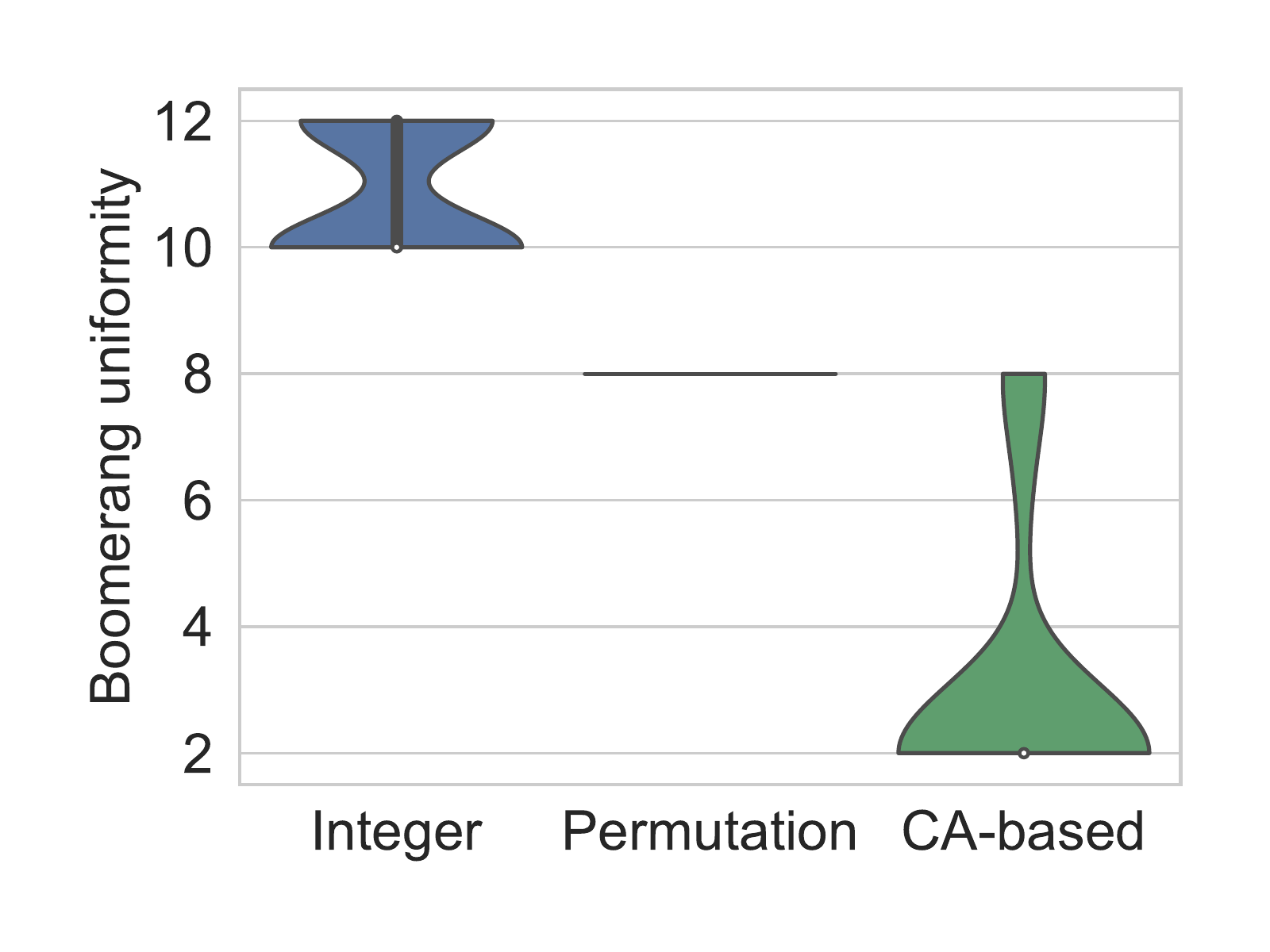}
         \caption{$5\times 5$}
         \label{fig:violin5}
     \end{subfigure}
     \hfill
     \begin{subfigure}[b]{0.49\textwidth}
         \centering
         \includegraphics[width=\textwidth]{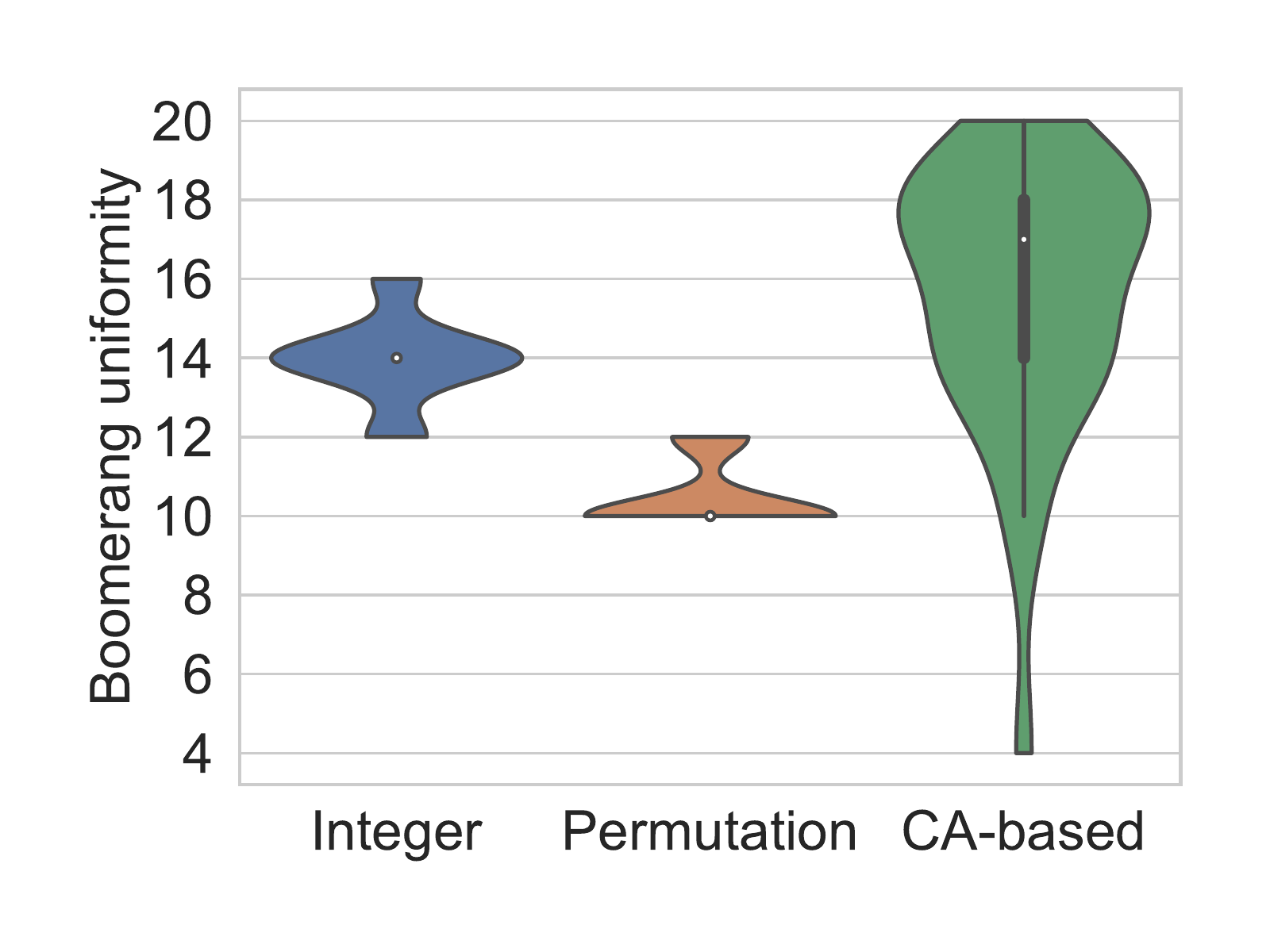}
         \caption{$6\times 6$}
         \label{fig:violin6}
     \end{subfigure}
          \begin{subfigure}[b]{0.49\textwidth}
         \centering
         \includegraphics[width=\textwidth]{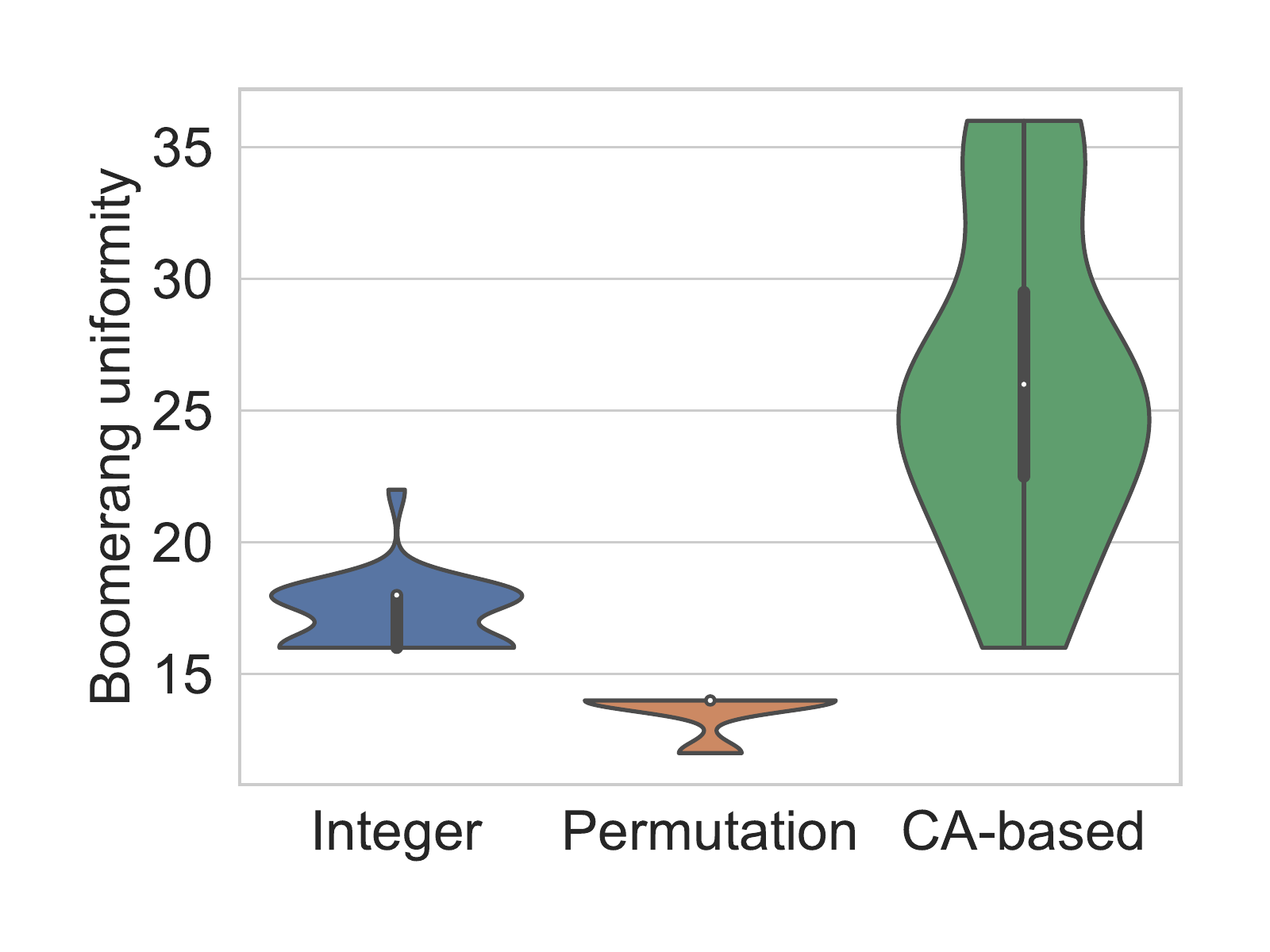}
         \caption{$7\times 7$}
         \label{fig:violin7}
     \end{subfigure}
          \begin{subfigure}[b]{0.49\textwidth}
         \centering
         \includegraphics[width=\textwidth]{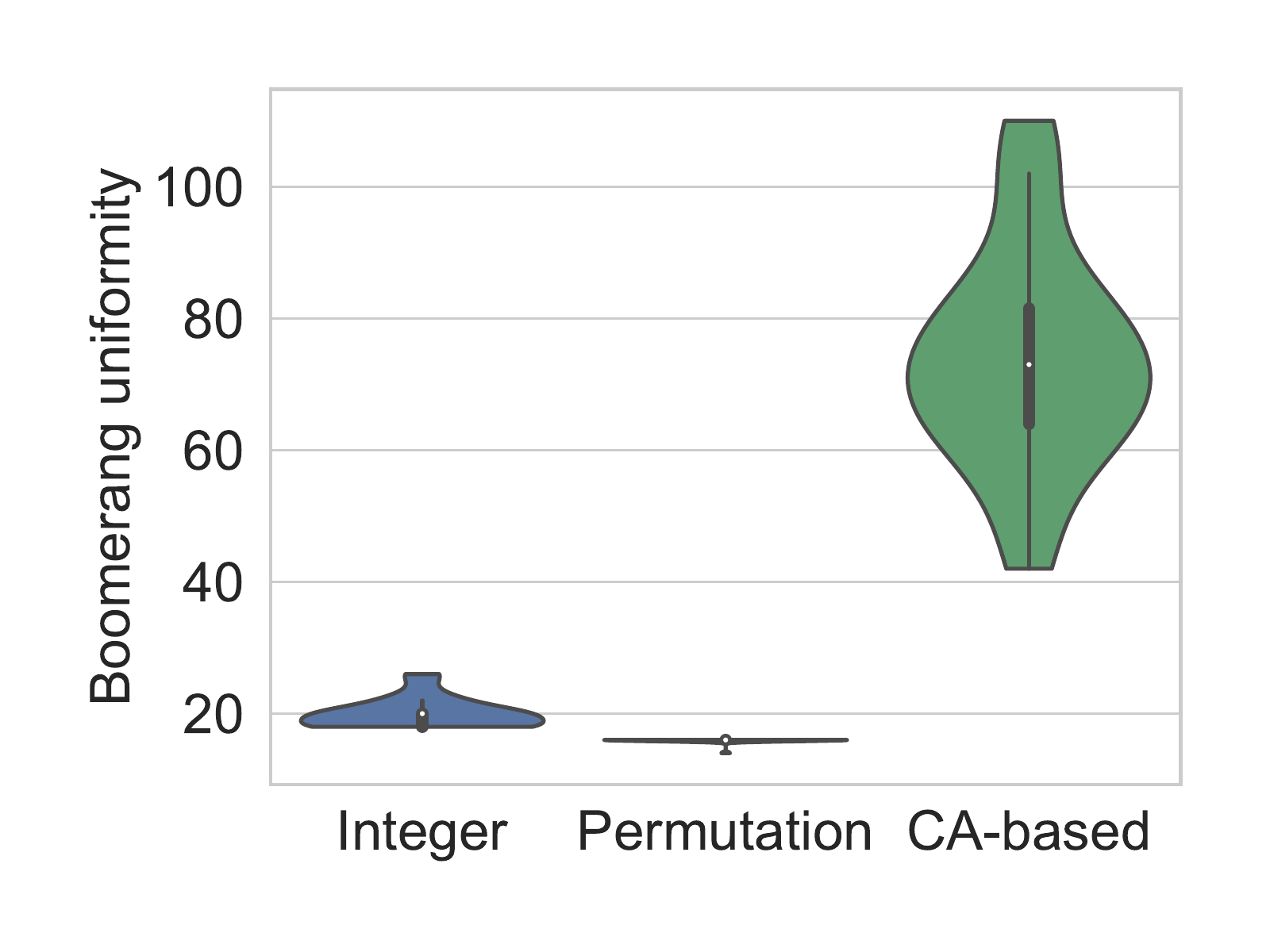}
         \caption{$8\times 8$}
         \label{fig:violin8}
     \end{subfigure}
        \caption{Violin plots outlining the distributions of the results obtained by each encoding.}
        \label{fig:violin}
\end{figure}

Figure~\ref{fig:violin} shows the violin plots of the results obtained with the evolutionary algorithm for each encoding and S-box size.
The distributions of the violin plots were cut off to better outline the minimum and maximum ranges that were obtained using each encoding. 
We can see that using the permutation encoding results in the lowest distribution of the results across all S-box sizes.
On the other hand, the evolutionary algorithm seems to be the least stable when using the CA-based representation since even in cases when it obtained the best overall result, the solutions from the 30 runs were quite dissipated.
As the size of the S-boxes increases, the dissipation of the CA-based results increases even further.
Finally, the integer encoding also obtains quite dispersed results, more dispersed than those of the permutation encoding, but the dispersion does not seem to increase as significantly with the increase of S-box sizes. 

Based on the summary of results in this section, we see that no single encoding leads to the best results for all S-box sizes. 
For smaller S-box sizes, the CA-based encoding is most appropriate, whereas, for larger S-box sizes, the permutation encoding achieves the best results.
On the other hand, the integer encoding only achieves the best result for the smallest S-box size.
Since it always achieves results worse than those of the permutation encoding, there is little benefit in using this encoding, especially as it does not provide any guarantee that it will obtain balanced S-boxes. 

In comparison with theoretical results (algebraic constructions), we can also provide some observations.
For $4\times 4$, all encodings with EA reach a boomerang uniformity of 6, which is the optimal value.
For $5\times 5$, we reach optimal boomerang uniformity with CA-based encoding, using both EA and random search.
For $6\times 6$, we reach optimal boomerang uniformity for a non-APN function (more precisely, differentially 4-uniform function). Further analysis shows this function is quadratic (algebraic degree equal to 2). It is known that such functions have optimal values of boomerang uniformity. 
Considering that for better boomerang uniformity we would need to find smaller differential uniformity (which, to the best of our knowledge, was never achieved with metaheuristics), we believe our result is excellent.
For $7\times 7$, we cannot reach the optimal value for boomerang uniformity. The best value is 12 with permutation encoding. It is known that the maximal value for differentially 4-uniform quadratic permutation is at most 4. Since there are known results with metaheuristics reaching APN functions~\cite{DBLP:journals/ccds/MariotPLJ19}, we can consider our result here rather poor, indicating that evolving for boomerang uniformity may well be more difficult than evolving for differential uniformity.
For $8\times 8$, the best results are far from optimal values achievable with algebraic constructions.

\begin{figure}[!ht]
     \centering
     \begin{subfigure}[b]{0.49\textwidth}
         \centering
         \includegraphics[width=\textwidth]{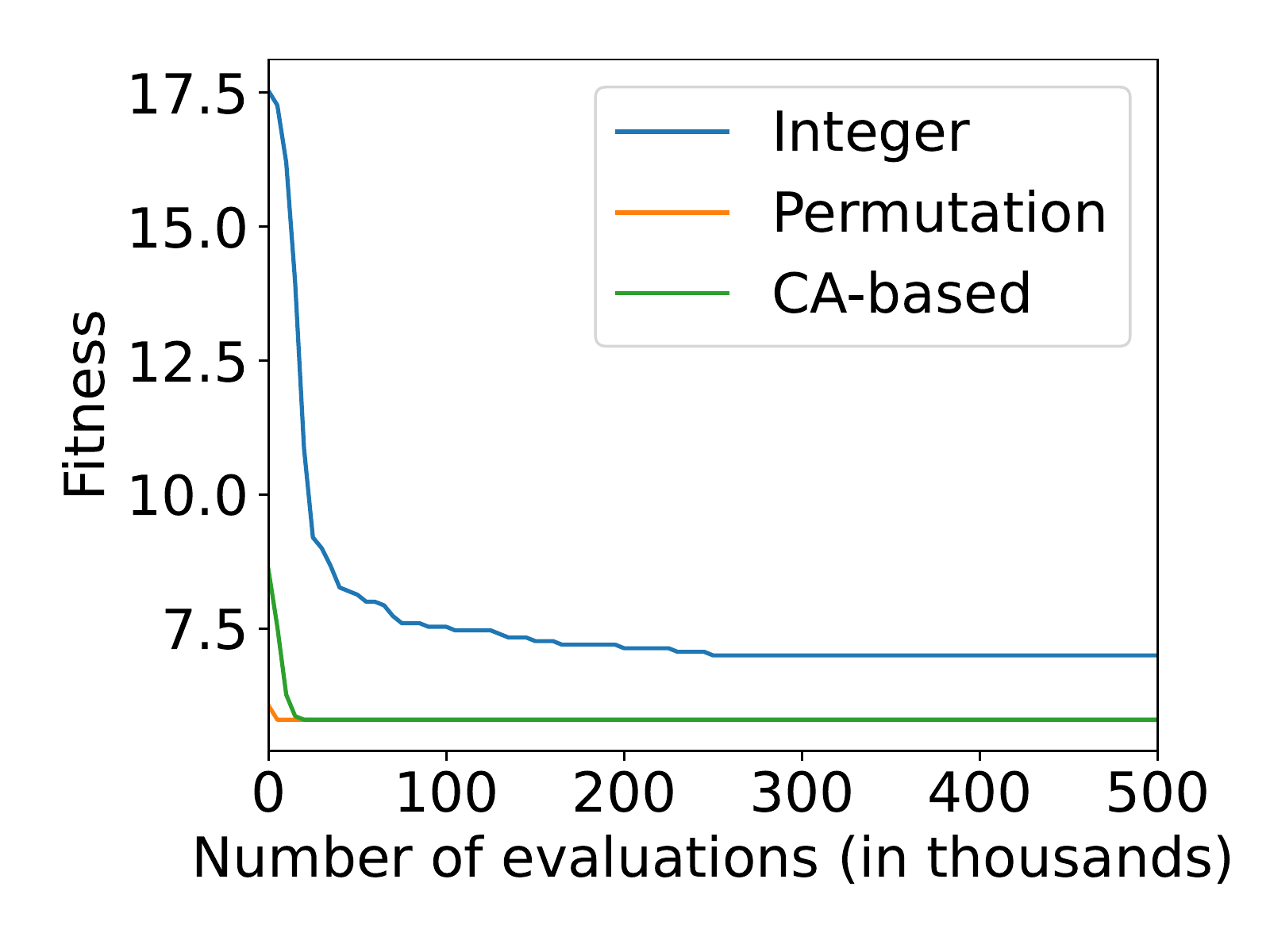}
         \caption{$4\times 4$}
         \label{fig:c4x4}
     \end{subfigure}
     \hfill
     \begin{subfigure}[b]{0.49\textwidth}
         \centering
         \includegraphics[width=\textwidth]{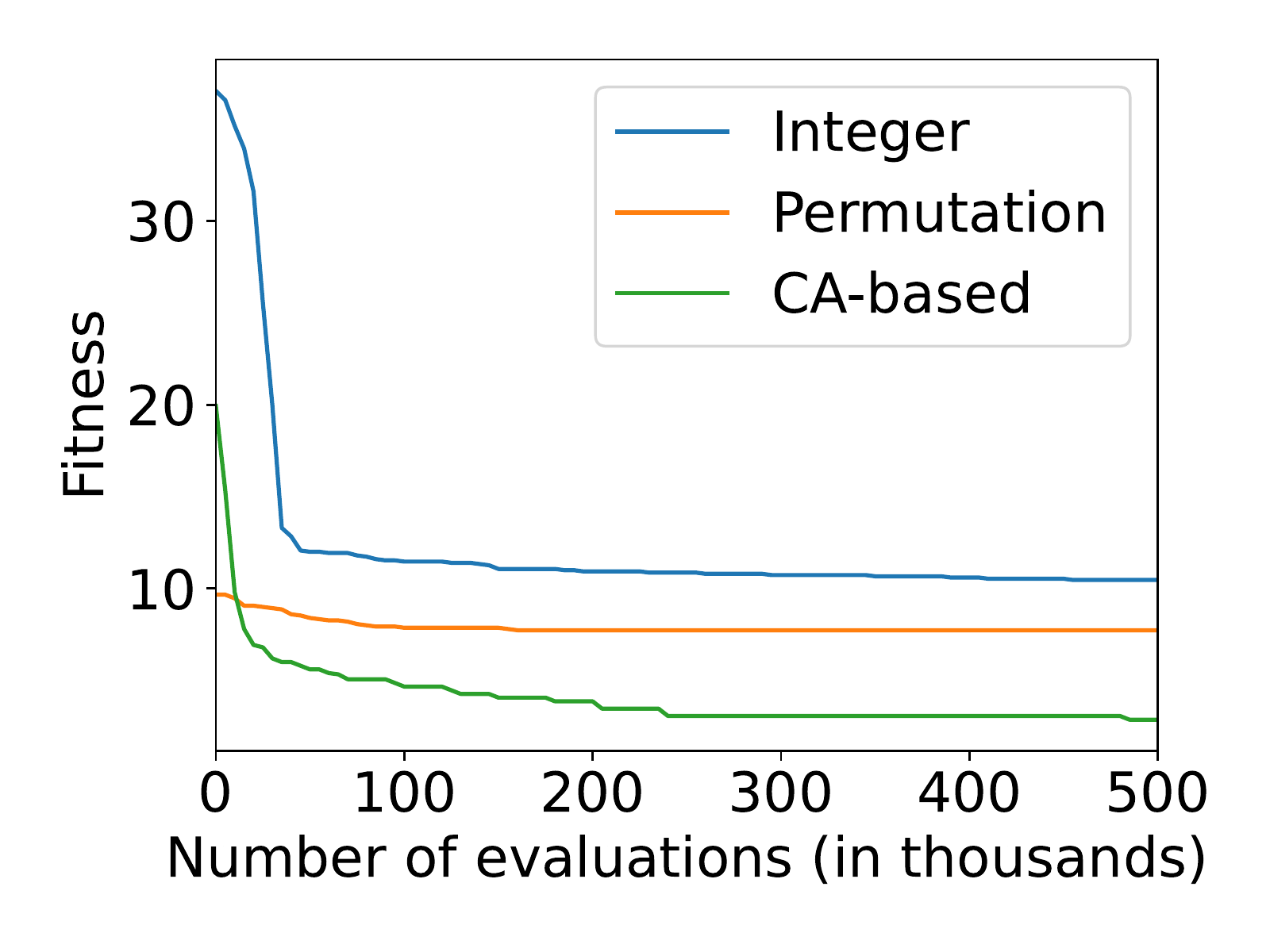}
         \caption{$5\times 5$}
         \label{fig:c5x5}
     \end{subfigure}
     \hfill   
     \begin{subfigure}[b]{0.49\textwidth}
         \centering 
         \includegraphics[width=\textwidth]{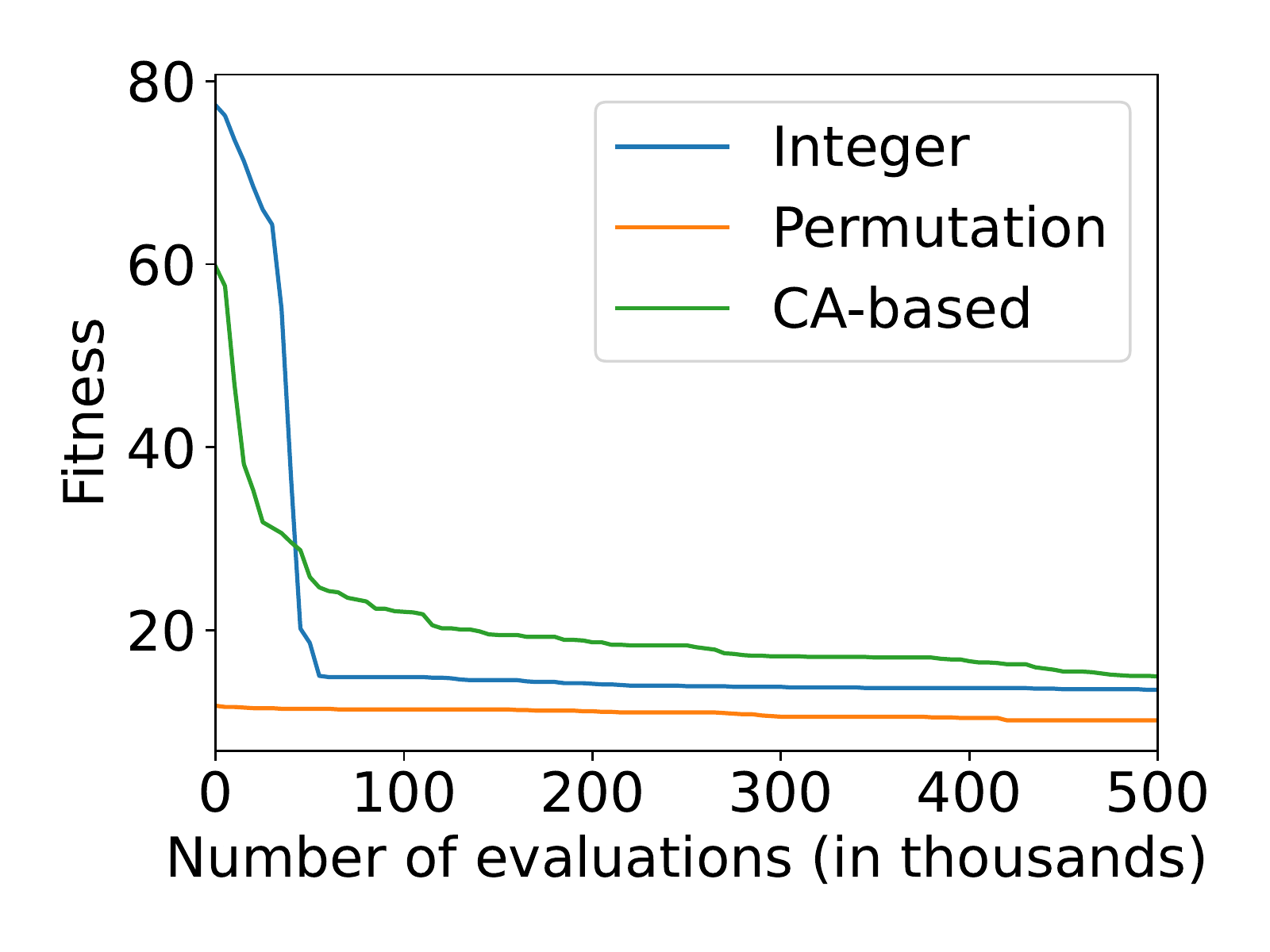}
         \caption{$6\times 6$}
         \label{fig:c6x6}
     \end{subfigure}
          \begin{subfigure}[b]{0.49\textwidth}
         \centering
         \includegraphics[width=\textwidth]{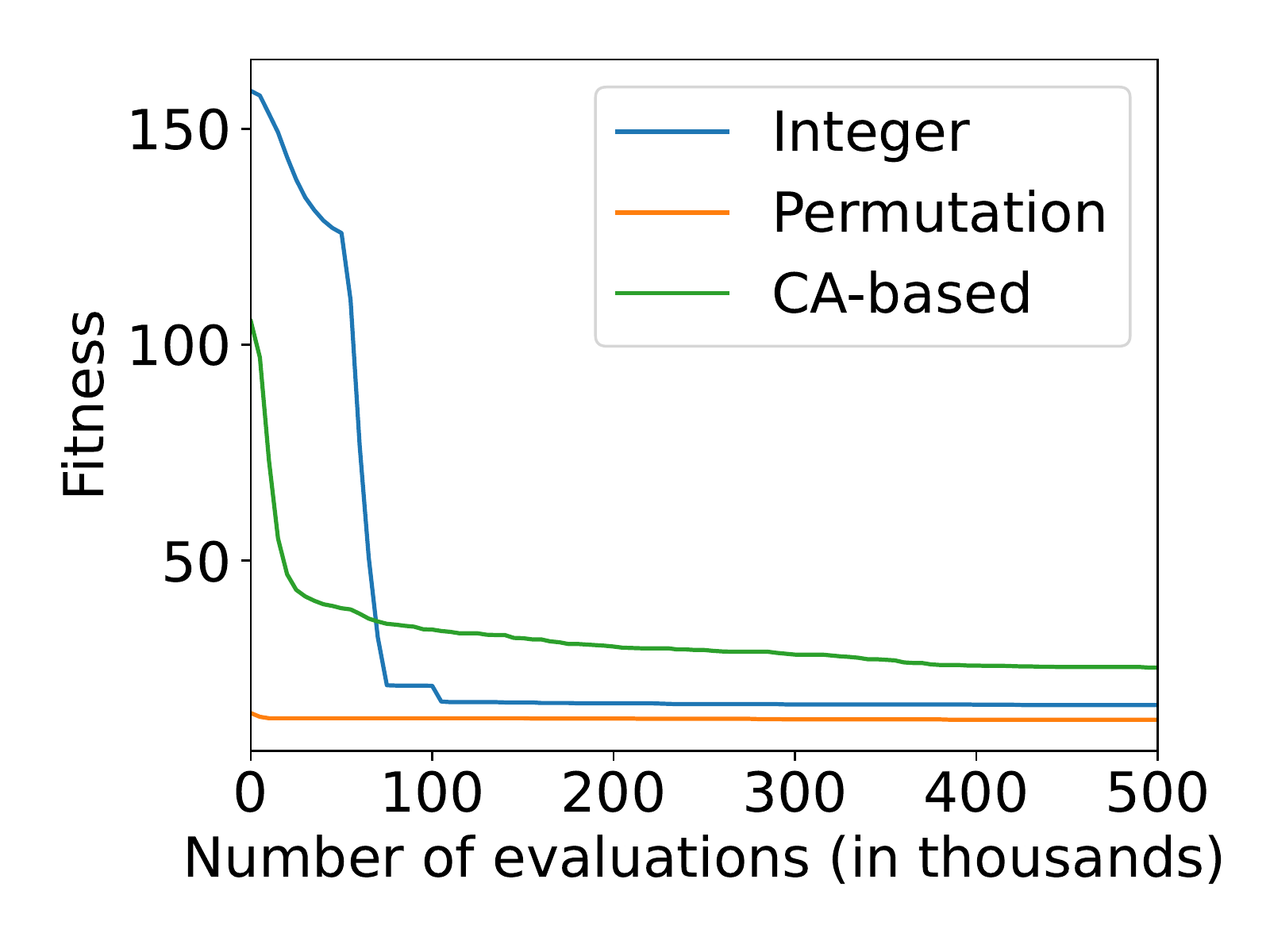}
         \caption{$7\times 7$}
         \label{fig:c7x7}
     \end{subfigure}
          \begin{subfigure}[b]{0.49\textwidth}
         \centering
         \includegraphics[width=\textwidth]{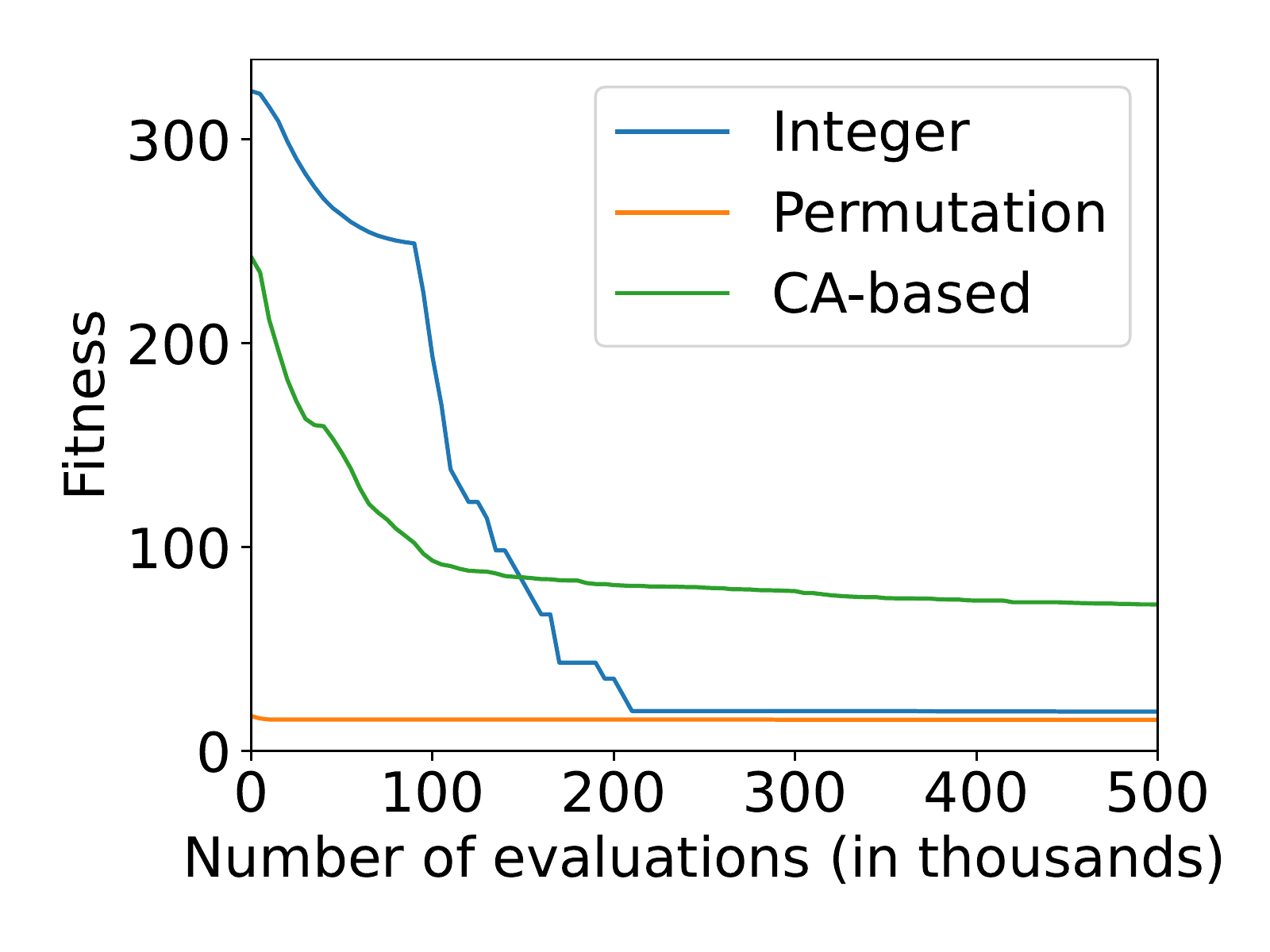}
         \caption{$8\times 8$}
         \label{fig:c8x8}
     \end{subfigure}
        \caption{Convergence patterns obtained for the tested encodings and S-box sizes.}
        \label{fig:conv}
\end{figure}

Figure~\ref{fig:conv} provides convergence graphs for all S-box sizes. Note how for $4\times 4$ size, permutation and CA-based encoding reach optimal fitness rather fast. 
For larger S-box sizes, the permutation encoding starts with much better initial solutions but improves them only slightly during the evolution process. The integer encoding demonstrates a slower convergence, and after a certain number of evaluations, it can be observed that the solutions are not improved further.
Finally, the CA-based encoding has the slowest convergence for large S-box sizes. However, the figures show that fitness is still being improved, meaning that it would be possible to achieve better results given more time. 
Nevertheless, it is questionable how long and, if at all, the CA-based encoding could reach solutions of the same quality as those obtained by the other two encodings. 

\subsection{Multi-objective Optimization}

\begin{figure}[!ht]
     \centering
     \begin{subfigure}[b]{0.49\textwidth}
         \centering
         \includegraphics[width=\textwidth]{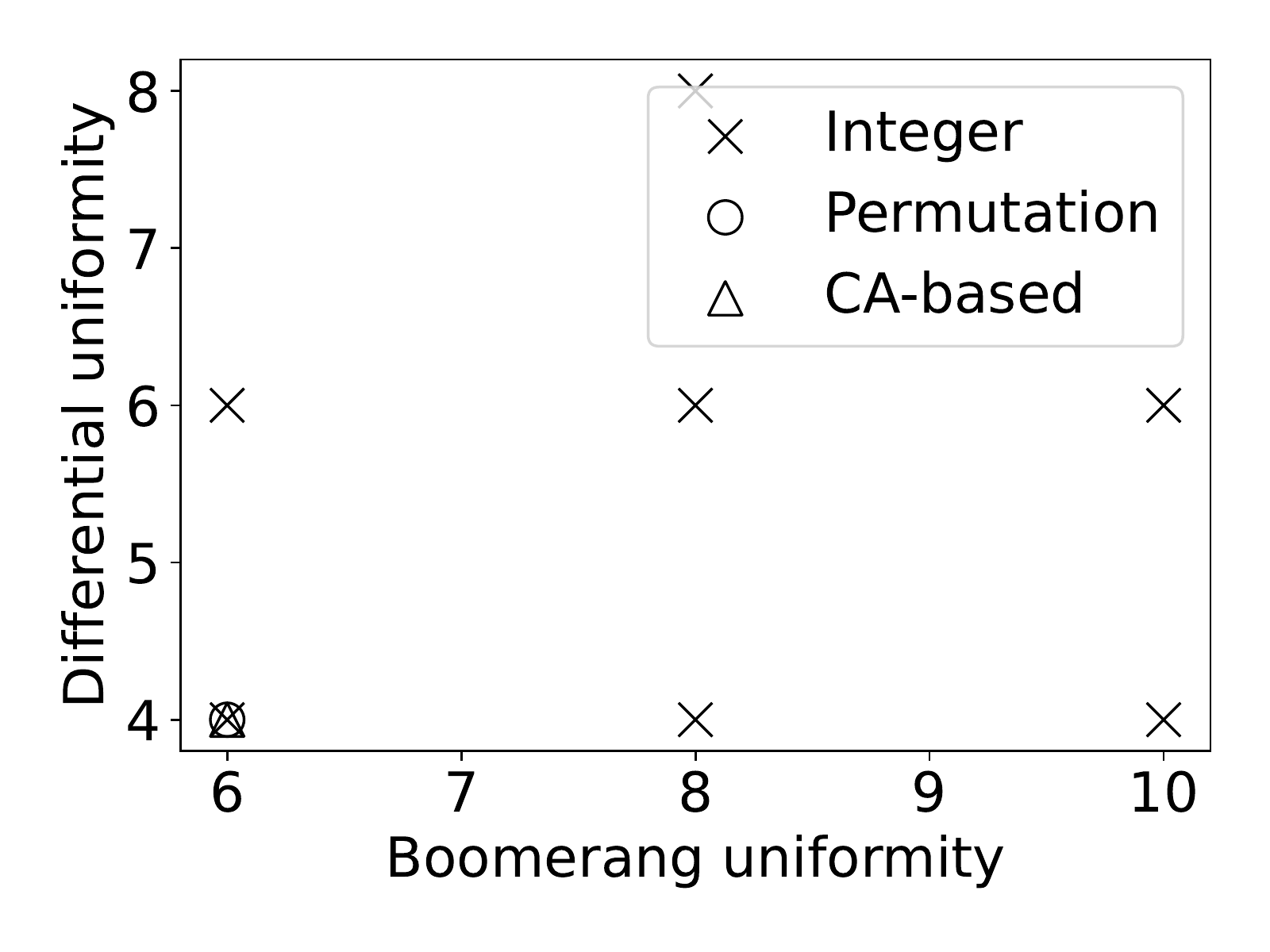}
         \caption{$4\times 4$}
         \label{fig:p4x4}
     \end{subfigure}
     \hfill
     \begin{subfigure}[b]{0.49\textwidth}
         \centering
         \includegraphics[width=\textwidth]{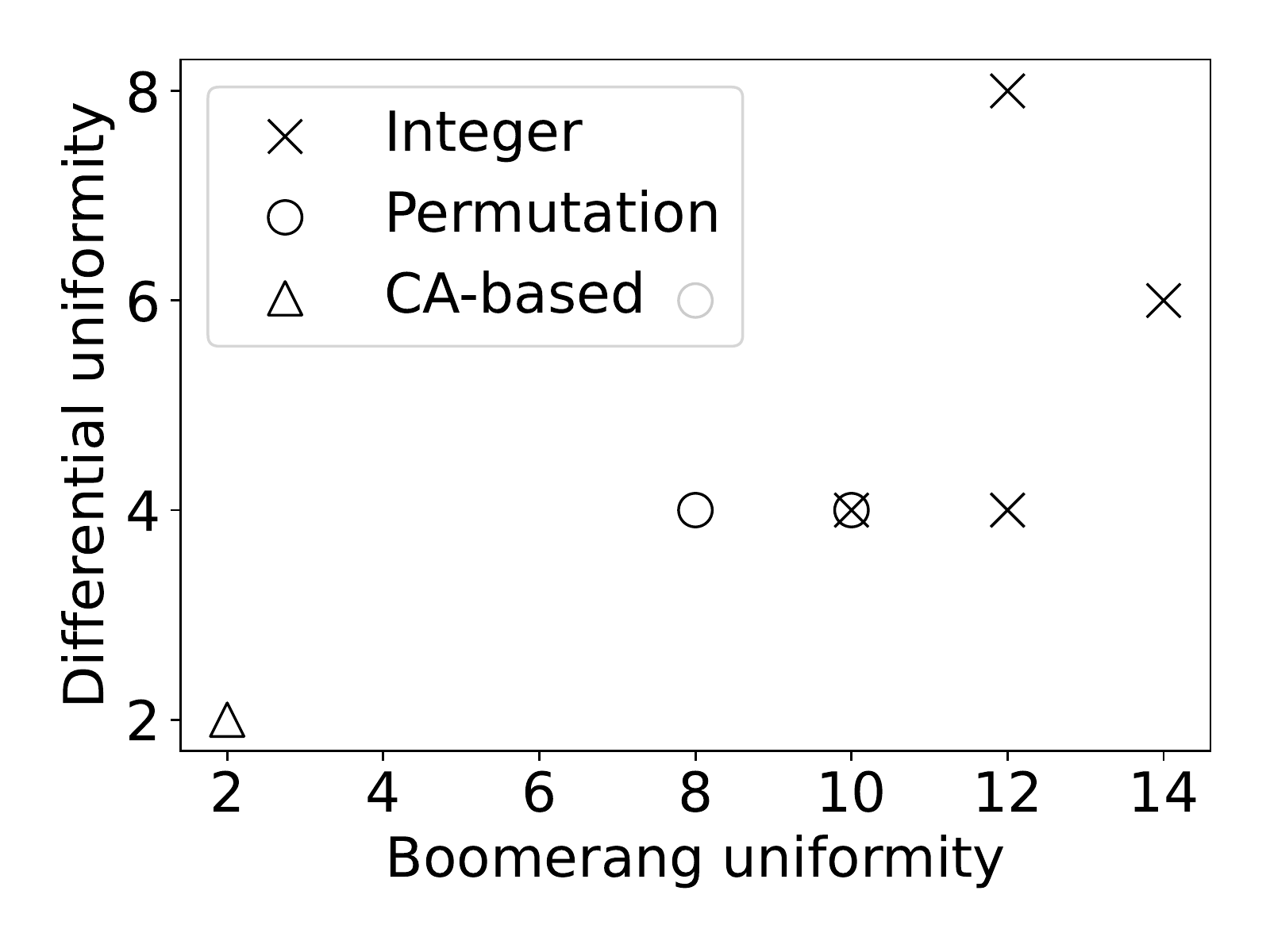}
         \caption{$5\times 5$}
         \label{fig:p5x5}
     \end{subfigure}
     \hfill
     \begin{subfigure}[b]{0.49\textwidth}
         \centering
         \includegraphics[width=\textwidth]{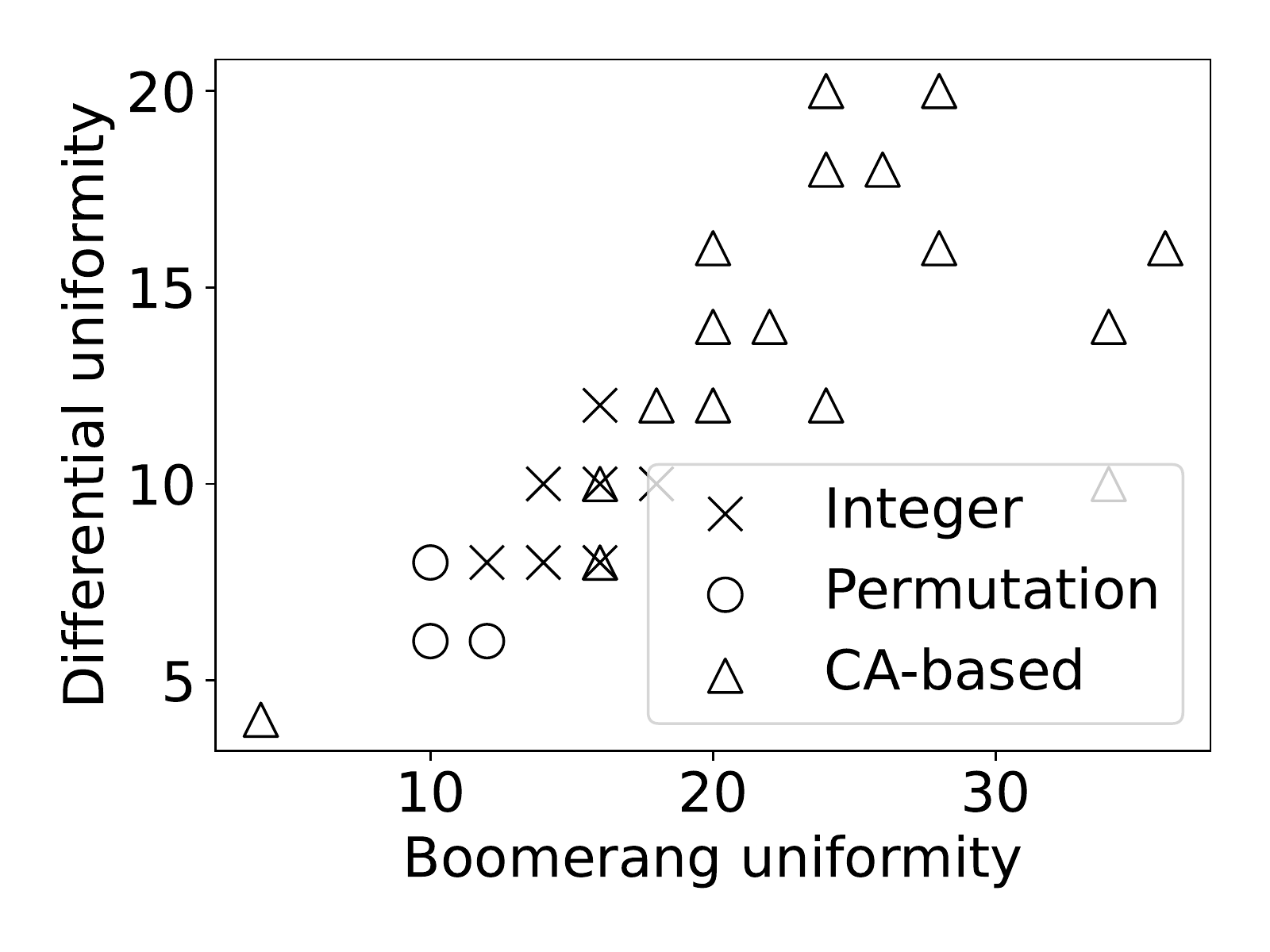}
         \caption{$6\times 6$}
         \label{fig:p6x6}
     \end{subfigure}
          \begin{subfigure}[b]{0.49\textwidth}
         \centering
         \includegraphics[width=\textwidth]{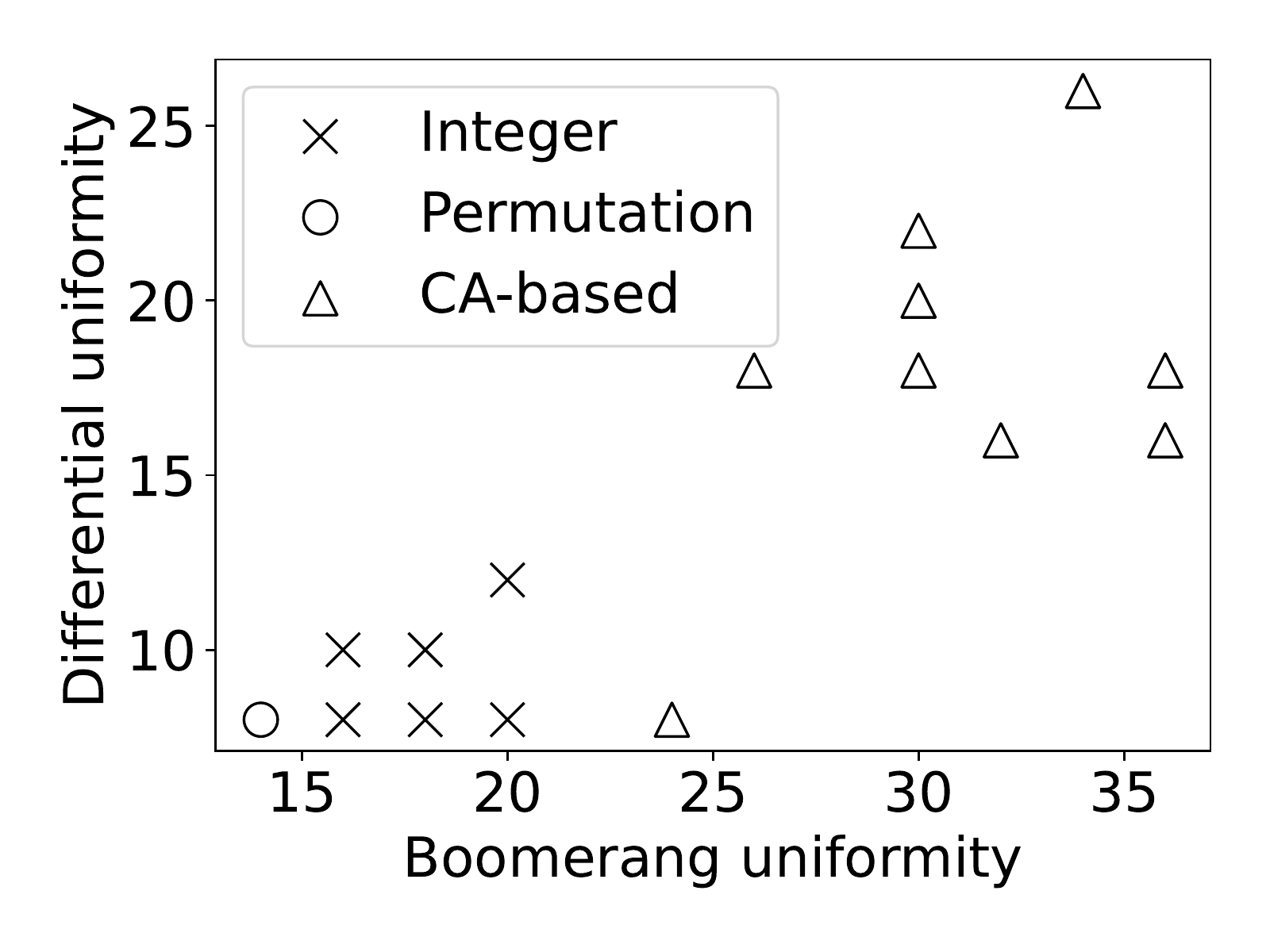}
         \caption{$7\times 7$}
         \label{fig:p7x7}
     \end{subfigure}
          \begin{subfigure}[b]{0.49\textwidth}
         \centering
         \includegraphics[width=\textwidth]{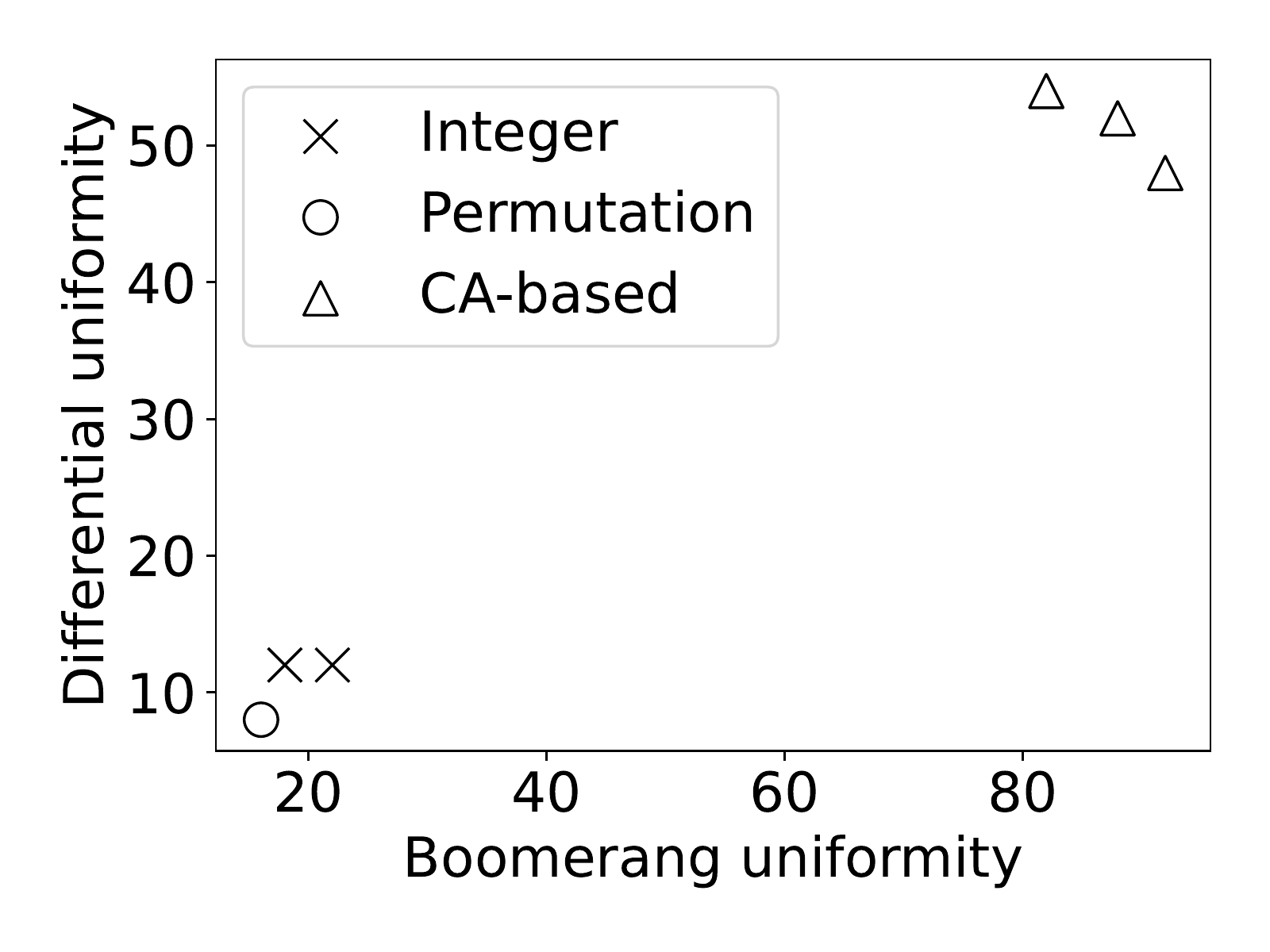}
         \caption{$8\times 8$}
         \label{fig:p8x8}
     \end{subfigure}
        \caption{Pareto fronts obtained for simultaneous optimization of boomerang and differential uniformity.}
        \label{fig:pareto}
\end{figure}

This section outlines the results obtained by simultaneously optimizing boomerang and differential uniformity using NSGA-II.
Figure~\ref{fig:pareto} shows the union of the Pareto fronts obtained from the 30 runs of NSGA-II for each encoding and S-box size. 
Since the figure denotes the union of all Pareto fronts, the outlined points do not necessarily form a single Pareto front.
This was done to illustrate better the distribution of the obtained solutions for each encoding. 

Even when considering multi-objective optimization, we see that the performance of the individual encodings follows a similar pattern as the one observed for the single-objective case.
For the S-boxes of size $4\times 4$, all encodings obtained the best possible result.
%However, this time the CA-based encoding does not obtain the best result for the S-box size of $4\times 4$, but rather the best results are obtained by the remaining two encodings. 
The permutation and CA-based encodings are more stable, as each run obtains the best solution, whereas the integer encoding sometimes obtains much worse solutions, which can be seen from the larger dissipation of solutions in the objective space.
For sizes $5 \times 5$ and $6 \times 6$, the overall best result is obtained by the CA-based encoding.
For these sizes, the permutation encoding clearly demonstrates to be superior to the integer encoding as it obtains better solutions for both criteria.
For the S-box size $6\times 6$, we also observe that the performance of the CA-based encoding started to deteriorate significantly, as in many cases, it obtained quite poor solutions.
Thus, one could say that the best solution it obtained was more due to luck.
Finally, for the remaining two sizes, we clearly see that the permutation-based encoding achieves the best result, which dominates the results obtained by the other two.
%The CA-based encoding achieves the worst results, which just continue deteriorating as the size of the S-box increases. 

\section{Conclusions and Future Work}
\label{sec:conclusions}

This work is the first paper that considers the evolutionary algorithms' perspective in designing S-boxes with good values of boomerang uniformity.
We run experiments for the relevant S-box sizes and three different solution encodings, and we obtain rather interesting results. For sizes $4 \times 4$ up to $5\times 5$, we obtain optimal values for the boomerang uniformity. For $6\times 6$, we obtain optimal boomerang uniformity for non-APN functions.
Sizes larger than $6\times 6$ exhibit rather poor results, and Pareto fronts indicate it is even somewhat easier to obtain good differential uniformity than boomerang uniformity.

In future work, it would be interesting to consider optimizing for nonlinearity and boomerang uniformity, as the link between those two properties is even less understood~\cite{Mesnager2020}.
Moreover, it would be interesting to see how we can add additional information to the evolutionary search to improve the performance. For instance, it is known that differentially 4-uniform quadratic permutations have boomerang uniformity at most 12. Adding that to the evolutionary search could, on one side, make the search more efficient, but it would also make the fitness function more complex. Finally, from the optimization perspective, it would be interesting to consider metaheuristics other than evolutionary algorithms.

\bibliographystyle{abbrv}
\bibliography{bibliography}

\end{document}